# Discrimination, calibration, and point estimate accuracy of GRU-D-Weibull architecture for real-time individualized endpoint prediction

Dr. Xiaoyang Ruan [1]; Dr. Liwei Wang [1]; Dr. Michelle Mai [1]; Dr. Charat Thongprayoon [2]; Dr. Wisit Cheungpasitporn [2]; Dr. Hongfang Liu[1]*

[1] Department of Artificial Intelligence & Informatics, Mayo Clinic, Rochester, MN, United States

[2] Department of Internal Medicine, Mayo Clinic, Rochester, MN, United States

Corresponding Author:

Hongfang Liu, PhD

Department of Artificial Intelligence & Informatics, Mayo Clinic

200 First Street SW

Rochester, MN 55905

United States

Phone 1 507 293 0057

Email: liu.hongfang@mayo.edu

## Abstract

Real-time individual endpoint prediction has always been a challenging task but of great clinic utility for both patients and healthcare providers. With 6,879 chronic kidney disease stage 4 (CKD4) patients as a use case, we explored the feasibility and performance of gated recurrent units with decay that models Weibull probability density function (GRU-D-Weibull) as a semi-parametric longitudinal model for real-time individual endpoint prediction. GRU-D-Weibull has a maximum C-index of 0.77 at 4.3 years of follow-up, compared to 0.68 achieved by competing models. The L1-loss of GRU-D-Weibull is ~66% of XGB(AFT), ~60% of MTLR, and ~30% of AFT model at CKD4 index date. The average absolute L1-loss of GRU-D-Weibull is around one year, with a minimum of 40% Parkes' serious error after index date. GRU-D-Weibull is not calibrated and significantly underestimates true survival probability. Feature importance tests indicate blood pressure becomes increasingly important during follow-up, while eGFR and blood albumin are less important. Most continuous features have non-linear/parabola impact on predicted survival time, and the results are generally consistent with existing knowledge. GRU-D-Weibull as a semi-parametric temporal model shows advantages in built-in parameterization of missingness, native support for asynchronously arrived measurement, capability of output both probability and point estimates at arbitrary time point for arbitrary prediction horizon, improved discrimination and point estimate accuracy after incorporating newly arrived data. Further research on its performance with more comprehensive input features, in-process or post-process calibration are warranted to benefit CKD4 or alike terminally-ill patients.

**Background**

Individual endpoint prediction has always been an intriguing and challenging task. Models that can accurately predict at individual level the risk of reaching endpoint, or more specifically, a point estimate of time-to-endpoint are highly desired, since precise estimation of the condition of an individual patient can prompt proper medical intervention. This is especially meaningful for chronic kidney disease (CKD)[1–7] or similar scenarios that see increasing mortality rate [8–10] with high world-wide prevalence [11]. However, due to the nature of data sparsity, asynchronicity, individual uncertainty, and intrinsic large statistical variations, most existing predictive modeling studies belong to static models (i.e. unable to incorporate updated measurements) with binary/multi-class outputs on a fixed prediction horizon. To name a few, take CKD for example, Kusiak et al. [13] used decision tree algorithm and rough-set algorithm to achieve individual level above/below median survival time prediction of hemodialysis patients. Tangri [14] modeled the progression of CKD to kidney failure using the Cox proportional hazard model with static risk scoring equation. Noia et al. [15] created an ensemble of 10 feed forward networks (through 10-fold cross-validation) trained with a large cohort to predict ESRD risk as a binary outcome of IgA nephropathy patients. Zhang et al. [16] used a multi-layer perceptron to predict CKD 5-year survival as a binary outcome. Xiao et al. [17] predicted CKD severity status using variables available through regular blood tests instead of 24-h urinary protein. Naqvi et al. [18] predicted survival status as binary outcome at 3 timepoints of kidney graft patients by several commonly used machine learning mechanisms. Zou et al. [19] and Lee et al. [20] used machine learning with static input features to predict risk of ESRD in diabetic kidney disease and sepsis survivors, respectively. Yuan et al. [21] Reviewed the application of AI technology in alerting, diagnostic assistance, treatment guidance and prognosis evaluation of acute and chronic kidney disease. While static models with binary/multi-class classifiers may have state-of-art prediction accuracy, they lose flexibility that comes from directly modeling time duration to events: healthcare practitioners are forced to decide predetermined duration(s) where an event is to occur or not. Moreover, the models become less applicable, if not useless, when medical interventions change the trajectory of disease progression in real time. It is also nearly impossible to construct and maintain separate survival models at different time points of disease progression due to the fast-changing nature of the diseases, the labor involved in sample collection, as well as the sparsity and missingness of measurements, which are easily anticipated during long time follow-up.

It is thus tempting to have a model that can dynamically update risk assessment while being less demanding on data integrity. We envision gated recurrent units with decay that models Weibull probability density function (GRU-D-Weibull) as a such potential candidate, for its ability to take longitudinal input, automated missing parameterization, and output individualized survival distribution at arbitrary time points. To this end, we systematically compared GRU-D-Weibull with several more traditional survival prediction models including accelerated failure time (AFT) model , XGBoost AFT, and multi-target logistic regression (MTLR) model on a rich panel of performance metrics. Our results demonstrated the capability of GRU-D-Weibull in achieving higher concordance rate and better point estimate accuracy, and pave the road for future improvements.

**Materials and Methods**
The study includes a total of 9,479 patients with CKD4 diagnosis (ICD9 code 585.4 or ICD10 code N18.4) between 2005 and end of 2017 at Mayo Clinic Rochester. We excluded 1,862 patients with any kidney transplant record, or with CKD5 or end stage renal disease (ESRD) diagnosis outside of Rochester (to minimize the impact of lost follow-up or very sparse lab measurement of migrating patients). After that 738 patients with no lab record were removed. After exclusion, a total of 6,879 patients were included in the final analysis (Fig 1), which has baseline population characteristics shown in Table 1. Specifically, we use a composite endpoint defined as CKD5, ESRD, or death, whichever occurs first. While the term "survival" usually means "time to death", throughout the paper we use "survival" interchangeably with "time to composite endpoint" for simplicity. The CKD4 diagnosis date distribution is shown in Fig 1.

**5-fold cross validation and held-out dataset**
From the 6879 qualified patients, 1879 were selected through a completely random mechanism as a held-out dataset. For the remaining 5000 patients, censored stratification strategy (put approximately the same number of uncensored observations in each fold but not pay any attention to event time) was used to split the patients into 5 chunks (n=1000 per chunk). Notably, once the sample indices for the held-out and 5-fold datasets are determined, these indices are used consistently across all models (i.e. same indices for AFT, XGB(AFT), MTLR, and GRU-D-Weibull model) for fair comparison. The detailed training, validation, testing, and held-out application strategies are illustrated in Sup Fig 6.

| Table 1 Baseline characteristic of study population | |
|---|---:|
| **Number of patients** | 6879 |
| **Age at CKD4 diagnosis** | 75(65,83) |
| **Gender** | |
| Male | 3743(54%) |
| Female | 3136(46%) |
| **Race** | |
| Caucasian | 6231(91%) |
| Non-caucasian | 648(9%) |
| **Follow-up time(years)** | 2.2(0.5,4.8) |
| **Censoring rate** | 3370(49%) |
| **Events type \*** | |
| Death | 2663(76%) |
| CKD5 | 509(15%) |
| ESRD | 510(15%) |
| Data shown as n(%) or median(IQR) | |
| \* 169 patients have CKD5 and ESRD on the same day. 2 patients have ESRD and death on the same day. 2 patients have CKD5 and death on the same day | |

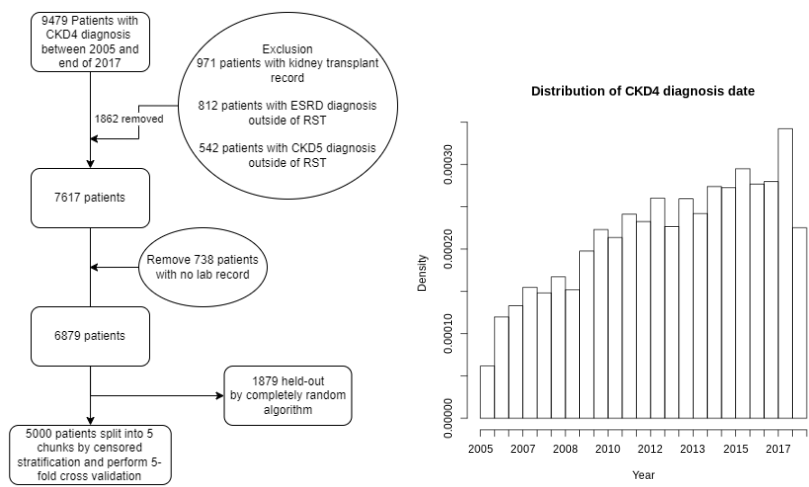

**Fig 1** Sample preprocessing workflow and distribution of CKD4 diagnosis date

## GRU-D model architectures

The basic architecture of the GRU-D model has been systematically described by [22], here we only recapitulate the equations for handling missing values.

$$\hat{x}_t^d = m_t^d x_t^d + (1 - m_t^d)(\gamma_{x_t}^d x_{t'}^d + (1 - \gamma_{x_t}^d)\tilde{x}^d)$$

where $m_t^d$ is the missing value indicator for feature $d$ at timestep $t$. $m_t^d$ takes value 1 when $x_t^d$ is observed, or 0 otherwise, in which case the function resorts to weighted sum of the last observed value $x_{t'}^d$ and empirical mean $\tilde{x}^d$ calculated from the training data for the $d$th feature. Furthermore, the weighting factor $\gamma_{x_t}^d$ is determined by

$$\gamma_t = exp\{-max(0, W_\gamma \delta_t + b_\gamma)\}$$

where $W_\gamma$ is a trainable weights matrix and $\delta_t$ is the time interval from the last observation to the current timestep. When $\delta_t$ is large (i.e. the last observation is far away from current timestep), $\gamma_t$ is small, results in smaller weights on the last observed value $x_{t'}^d$, and higher weights on the empirical mean $\tilde{x}^d$ (i.e. decay to mean).

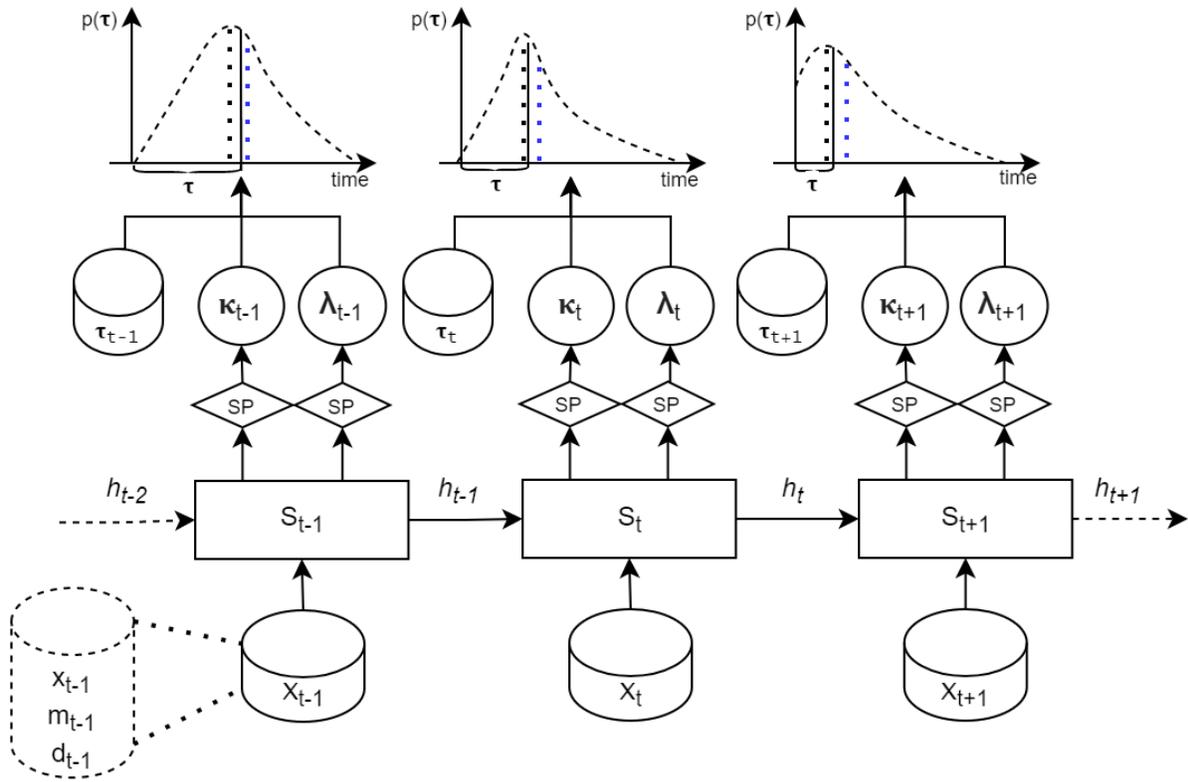

**Fig 2** GRU-D architecture that outputs 2-parameter Weibull's distribution. The shape parameter $\kappa$ and scale parameter $\lambda$ are updated at each timestep to maximize composite loss of negative log likelihood Weibull's PDF plus MSLE between predicted median survival time (PMST) and observed remaining survival time $TR$. In the PDF plots, vertical dashed black and blue lines represent the mode and median, respectively, of Weibull's PDF, which changes over time.

$S_t$:  State of GRU-D cell at timestep t
$h_t$:  hidden output at timestep t
$SP$:  SoftPlus activation
$\tau$:  Time remaining to event (or censoring)
$p(\tau)$:  Probability of reaching end point at time $\tau$
$\kappa$:  Shape parameter
$\lambda$:  Scale parameter
$x$:  Measurement value
$m$:  Missing indicator
$d$:  Delta time

In this study we used GRU-D architecture that models 2-parameter Weibull distribution (GRU-D-Weibull) (Fig 2). Specifically, the model is trained to minimize a composite loss function (equation (1)) composed of a) negative log likelihood of Weibull's PDF (*neglog*) and b) mean square log error of the difference between predicted median survival time (PMST) and observed remaining time to event (*MSLE*).

$$loss_{total} = \sum_{i=1}^{N} (w_i \sum_{t=1}^{T_i} (-1 * log(f(\tau_{it}, \lambda_{it}, \kappa_{it}))) + MSLE(\tau_{it}, Median(\hat{\tau}_{it}))) \quad (1)$$

$$loss_{mean} = loss_{total} / \sum_{i=1}^{N} T_i$$

where (subscripts $i$ and $t$ not displayed for clarity)

$$f(\tau; \lambda; \kappa) = h(\tau)S(\tau) = \frac{\kappa}{\lambda} \left(\frac{\tau}{\lambda}\right)^{\kappa-1} e^{-(\tau/\lambda)^{\kappa}} \quad (2)$$

$$MSLE(\tau, Median(\hat{\tau})) = (log(\tau+1) - log(\lambda * (-1 * log(0.5))^{1/\kappa} + 1))^2$$

Specifically, $N$ is the number of patients in a training batch, $T_i$ is the timesteps before event or censoring for patient $i$ (i.e. no contribution to the model if the patient developed an endpoint or is censored). $w_i$ is individualized weights.

$$w_i = \begin{cases} 1, & \text{if not censored.} \\ t_c/5, & \text{otherwise, time of censoring divide by 5(yrs).} \end{cases}$$

$Median(\hat{\tau})$ is the PMST. $\tau_{it}$ is the remaining time to event for uncensored patients or "Best Guess" survival time estimated by MTLR (equation (3)) for censored patients at timestep $t$.

$$BG(c_i) = c_i + \frac{\int_{c_i}^{\infty} S(\tau)d\tau}{S(c_i)} == c_i + \frac{\lambda \frac{1}{\kappa} \Gamma(\frac{1}{\kappa}, (\frac{c_i}{\lambda})^{\kappa})}{exp[-(\frac{c_i}{\lambda})^{\kappa}]} \quad (3)$$

Where $c_i$ is the censoring time for patient $i$. $S(c_i)$ is the probability of survival at time of censoring. $\lambda$ is approximated by mean survival time of MTLR. $\kappa$ is assumed to equal 1 (i.e. exponential distribution), since MTLR does not explicitly estimate a shape parameter. As we noticed that $BG(c_i)$ leads to overestimation of survival time when compared with uncensored patients (Sup Fig 4), we truncated $BG(c_i)$ to a maximum of 5 years to avoid overwhelming the network with unreasonably large guess values. Note $\tau$ decreases over time.

$h(\tau)$ is the unitarized hazard (i.e. probability of developing event in a unit of time) at time $\tau$. $S(\tau)$ is the probability of surviving up to time $\tau$. The pseudocode for training GRU-D-Weibull model is shown in Algorithm 1.

---

**Algorithm 1:** Timestep sensitive composite loss based training of GRU-D-Weibull

---

Input (for patient $i$): $X_i = [x_i, m_i, d_i]$
  $x_i, m_i, d_i$ respectively are $f$ by $T_i$ tensors represent $f$ features with $T_i$ timesteps
Intermediate output: $\{[\kappa_{i0}, \lambda_{i0}]..[\kappa_{iT_i}, \lambda_{iT_i}]\}$
Final output: Average loss for a training batch

Initialize batch loss $Loss_{batch}$ to an empty array
**for** each patient $i$ in a training batch **do**
  Prepare weight for patient $i$ as $w_i = 1$ if not censored, or
    $w_i = Time\_of\_censoring/5(yrs)$
  **for** $t = \{1..T_i\}$ **do**
    $\kappa_{it}, \lambda_{it}$ = GRU-D($x_{it}, m_{it}, d_{it}$)
    Prepare $\tau_{it}$ as the remaining time to event (uncensored)
      or best guess survival time (censored) for patient $i$ at timestep $t$
    $Loss_{it}$ = COMPOSITE-LOSS($\kappa_{it}, \lambda_{it}, \tau_{it}$)
    $WeightedLoss_{it} = w_i * Loss_{it}$
    Append($Loss_{batch}, WeightedLoss_{it}$)
  **end for**
**end for**
Calculate average batch loss $AvgLoss_{batch} = Average(Loss_{batch})$

Calculate model delta weight change BACKWARD($AvgLoss_{batch}$)
Use gradient descent to update model parameters $W = W + \eta W$

---

We emphasize that given observed remaining time to event $\tau_t$ at timestep $t$, any single part of the composite loss function (i.e. either *neglog* or *MSLE*) allows for theoretically infinite combinations of $\kappa$ and $\lambda$. This is illustrated in Fig 3 by letting $\tau = 3(yrs)$, where the combinations satisfying mode at

$$\tau = \lambda(\frac{\kappa - 1}{\kappa})^{1/\kappa}$$

for *neglog* are shown as black line, and the combinations minimizing *MSLE*

$$\tau = \lambda(-1 * log(0.5))^{1/\kappa}$$

are shown as blue line. By using a composite loss function, the choices of $\kappa$ and $\lambda$ are theoretically confined to a point (intersect at $\kappa = 3.25$). In practice, both $\kappa$ and $\lambda$ are spreaded within a smaller range under composite loss function (Sup Fig 1).

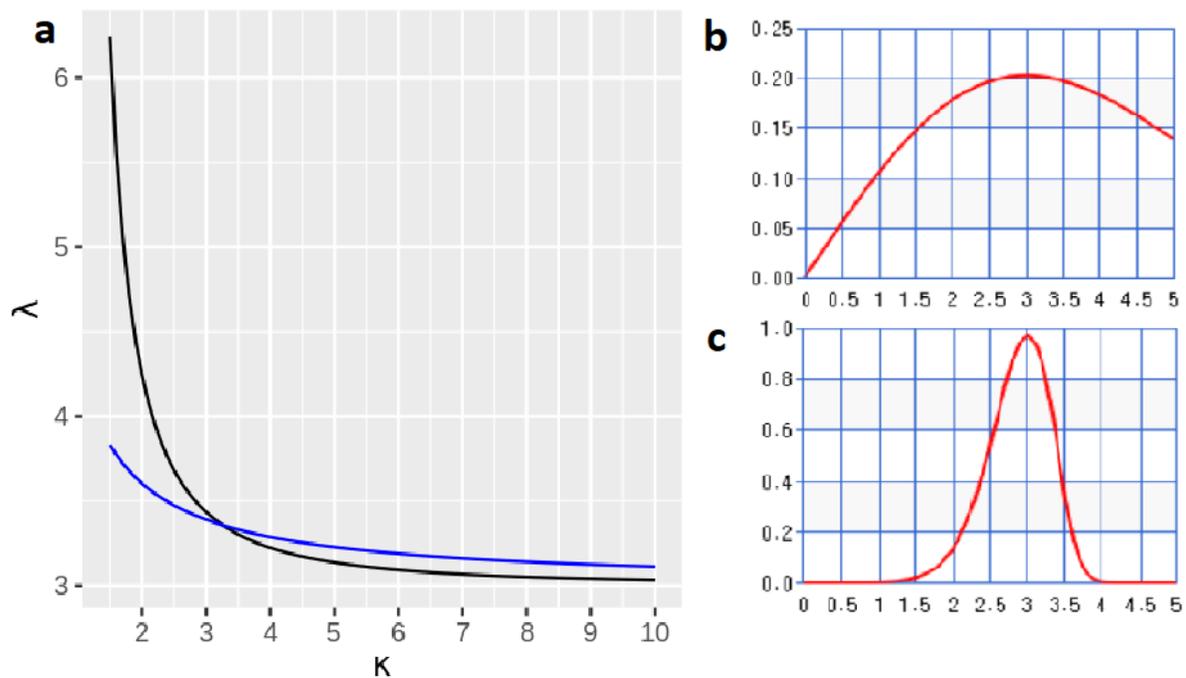

**Fig 3** a) Theoretical combinations of $\kappa$ and $\lambda$ that minimize *neglog* (black line), and *MSLE* (blue line) when the remaining time to event is 3 years. b) Weibull's PDF when $\kappa = 2$ and $\lambda = 4.24$, c) $\kappa = 8$ and $\lambda = 3.0$. Note both b) and c) have mode at 3.

Moreover, In a test run of GRU-D-Weibull models trained with only *neglog* or only *MSLE* loss function (Sup Fig 2), both scenarios showed remarkably different C-index across prediction horizons, due to a large variation in shape parameter $\kappa$ (Sup Fig 1).

**Hyperparameters and training schemes**

The hyperparameters and training schemes are determined manually by varying combinations of neuron size (20-100), learning rate (0.001-0.1), batch size (30-500), drop out ratio (0-0.5), optimizer (SGD, RMSprop, ADAM), clipnorm (1,3), clipvalue (1,3) and early stopping threshold that optimize the performance of GRU-D-Weibull on 5000 samples (1879 held-out not touched). Finally, we found that a combination of 40 neurons, learning rate 0.001, 50 epochs, batch size 500, max 4% validation/training loss difference, and ADAM optimizer with amsgrad has overall better performance in terms of training speed and stability (e.g. resistance to gradient explosion). According to our tests, while the choices of neuron size and learning rate are relatively flexible, a large batch size is necessary to avoid gradient explosion (probably due to the sparsity of certain patient data, which may easily deviate the error surface of the network), and at least 30-40 epochs are required.

### Input features and sampling schemes

A total of 19 features including 17 dynamic (i.e. changes over time) and 2 static features were used in current analysis (Table 2). eGFR was estimated through the CKD-EPI equation [23]. For the AFT, XGB(AFT) and MTLR models, dynamic features excluding comorbidity were based on the closest measurement within 1 year before or after CKD4 diagnosis. Comorbidities were based on ICD9 and ICD10 code within 10 years before CKD4 diagnosis.

For the GRU-D-Weibull model, a non-uniformly distributed sampling scheme (every 15 days within half year around CKD4 diagnosis, and otherwise every 30 days) was adopted for dynamic features, based on the distribution of time intervals between measurements (Sup Fig 7). This generated a total of 110 timesteps for each patient from 3 years before and up to 5 years after CKD4 diagnosis. Static features were replicated through timesteps. Continuous features like eGFR, SBP, BMI were z-score transformed based on mean and standard deviation (SD) calculated from training dataset. Categorical variables like comorbidity and gender were binary encoded. We assume a 100-day effect window once a comorbidity is identified, after which the effect is automatically handled by the missing value decay mechanism.

**Table 2 Features overview and preprocessing methods**

|  | Preprocessing | Feature list |
|---|---|---|
| **Dynamic features** | | |
| Lab (numeric) | z-score | Estimated Glomerular Filtration Rate (eGFR), Blood Albumin, Phosphorus, Calcium, urine Albumin-to-Creatinine ratio (uACR), Bicarbonate |
| Vital (numeric) | z-score | Systolic Blood Pressure (SBP), Dystolic Blood Pressure (DBP) |
| Comorbidty (binary) | binary | Diabetes Mellitus (DM), Congestive Heart Failure (CHF), Chronic Artery Disease (CAD), Cirrhosis, Dyslipidemia |
| Others | see detail | Age [divide by 100], Smoking [binary], Alcohol [binary], Body Mass Index (BMI) [z-score] |
| **Static features** | | |
| Demographics | binary | Gender, Race |

### Permutation feature importance across time

Permutation feature importance was performed in the held-out dataset by randomly shuffling the sample id of each feature. For each feature, C-index at each time point was evaluated before and after permutation and the difference (before - after) was considered as the dependence of the model on the corresponding feature. 5 permutations were performed on each of the 5 models trained through 5-fold cross validation. This produced 25 data points for each feature at each time point, from which mean value and confidence interval are derived.

**Partial dependence analysis**

Partial dependence analysis was performed in the held-out dataset by shifting the original value of corresponding feature, while keeping other features untouched. Specifically, age was shifted by -20 to +20 years. The features requiring z-score transformation were shifted by a z-score amount from -2 to 2 stepped by 0.5. For binary features, an all-zero version and a yearly diagnosis version were generated and compared with the original value. The shifting was performed independently for the 5 models obtained through 5-fold cross validation and PMST was averaged over all held-out samples on all 5 folds.

**Accelerated Failure Time (AFT) model**

AFT model assumed linear contribution of input variables to the link function that models the survival time $\tau$

$$ln(\tau) = x'\beta + \sigma * \epsilon$$

where $\beta$ is the regression coefficient. $\sigma$ is the scaling factor for $\epsilon$, which usually assume either normal, logistic, or extreme distribution (equation (4))

$$f(\epsilon) = exp(\epsilon)exp(-exp(\epsilon)) \quad (4)$$

In our case, extreme distribution is selected, since it leads to Weibull distribution (equation (2)) of the survival probability of time $\tau$. The CDF for probability of survival at time $\tau$ is given by

$$S(\tau) = exp[-(\frac{\tau}{exp(x'\beta)})^k] = exp[-(\frac{\tau}{\lambda})^k] \quad (\kappa = 1/\sigma)$$

and median survival time is defined as

$$Median(\tau) = \lambda(-1 * log(0.5))^{(1/k)}$$

Where $\lambda$ is scale and $\kappa$ is shape parameter for a 2-parameter Weibull distribution. Note since $\kappa$ is the same for all patients, AFT assumes constant proportional hazard between patients.

**XGBoost with AFT (XGB(AFT))**

XGBoost AFT is a gradient boosting version of AFT. It uses a decision tree ensemble of AFT models to approximate the observed survival time [24]

$$ln(\tau) = DecisionTree(x) + \sigma_{hyper} * \epsilon$$

Where $\tau$ is observed survival time. $\sigma$ is the scaling factor for $\epsilon$, which usually assumes either normal, logistic, or extreme distribution. Unlike the regular AFT model, which outputs an estimated scaling factor, the scaling factor $\sigma_{hyper}$ in XGBoost is a hyperparameter and must be manually specified. We chose $\sigma_{hyper} = 1$ for simplicity and the fact that $\tau$ then has a standard exponential distribution. A maximum tree depth of 6 and 300 rounds training were used, with early stopping on no improvement of validation dataset in 10 training rounds, which leads to about 200 rounds of training in practice.

**Multi-target logistic regression (MTLR) model**
MTLR jointly maximizes the probability of observed survival status at multiple time points [25]. For each patient the probability of observing $y_1, y_2, \cdots y_m$ (where m represents maximum follow-up time) is given by

$$P_\Theta(Y=(y_1, y_2, \ldots, y_m) \mid \vec{x}) = \frac{\exp(\sum_{i=1}^{m} y_i(\vec{\theta_i} \cdot \vec{x} + b_i))}{\sum_{k=0}^{m} \exp(f_\Theta(\vec{x}, k))},$$

Where $\Theta_i$ represents the coefficients vector at time point $i$, and

$$f_\Theta(\vec{x}, k) = \sum_{i=k+1}^{m}(\vec{\theta_i} \cdot \vec{x} + b_i)$$

In this study, we chose time points from integer 1 to 5 years. It is noteworthy that MTLR generates a separate vector of coefficients for each time point considered in the model, and does not have the assumption of proportional hazards across time.

**Imputation of input feature (for AFT, XGB(AFT), and MTLR models)**
Imputations were performed independently for 5-fold cross validation and holdout datasets using MICE package [26] with "predictive mean matching" algorithm. Three imputations were conducted for the 5-fold and the held-out dataset separately. The training and testing schemes are shown in Sup Fig 6

**Definition of evaluation metrics**
The definition of evaluation metrics including concordance rate, L1-loss, L1-margin loss, 1-Calibration, and Brier Score were systematically discussed by [27], and we briefly describe the metrics below. To emphasize, for both AFT and GRU-D-Weibull models, the estimated scale parameter $\lambda$ and shape parameter $\kappa$ are used to calculate the metrics. For the MTLR model, we directly use the package's built-in function to obtain predicted survival probability, mean/median survival time.

**Concordance index (C-index)**
The calculation of C-index is based on Harrell's definition [28]. C-indices were calculated separately for each imputed dataset. The pooled concordance and variance were obtained through taking the average of the concordance and variance obtained from 3 imputations.

**L1-loss, L1-margin loss**
For uncensored patients, L1-loss is defined as the average difference between PMST and observed event time (equation(5)). For censored patients, L1-margin loss is defined as the average difference between PMST and "Best Guess" event time (equation(6)).

$$L1 = \frac{1}{|V_u|} \sum_{i \in N_u} |\tau_i - Median(\hat{\tau}_i)| \qquad (5)$$

$$L1_{margin} = \frac{1}{|V_c|} \sum_{i \in N_c} |BG(c_i) - Median(\hat{\tau}_i)| \qquad (6)$$

where $V_u$ and $V_c$ are the subset of uncensored and censored patients, respectively. Both L1 and L1-margin loss were calculated for each imputation, and then averaged to obtain pooled mean prediction error.

**1-calibration**
1-calibration is measured by Hosmer-Lemeshow (HL) goodness-of-fit test of the difference between predicted and observed number of events, sorted by predicted survival probability into $B$ bins ($B = 10$ in our case). Equation(7) shows the calculation of HL score.

$$HL(V, \hat{S}(\tau^*)) = \sum_{b=1}^{B} \frac{(n_b(1 - KM_b(\tau^*)) - n_b\overline{p}_b)^2}{n_b\overline{p}_b(1 - \overline{p}_b)} \qquad (7)$$

where $V$ represents all censored and uncensored patients, $\tau^*$ is the prediction horizon, $n_b$ is the number of patients in a bin, $\overline{p}_b$ is the average predicted number of events in a bin at time $\tau^*$. $KM_b(\tau^*)$ is the observed probability of survival at time $\tau^*$. To calculate pooled HL value, $KM_b(\tau^*)$ and $\overline{p}_b$ were calculated for each imputed dataset separately and then averaged.

**Brier score**
Brier score is measured by weighting predicted probability of survival at prediction horizon $\tau^*$ against probability of censoring (equation (8)).

$$BS(V, \hat{S}(\tau^*)) = \frac{1}{|V|} \sum_{i=1}^{|V|} \left[ \frac{I[\tau_i \leq \tau^*, \delta_i = 1](\hat{S}(\tau^*|X_i))^2}{\hat{G}(\tau_i)} + \frac{I[\tau_i > \tau^*](1 - \hat{S}(\tau^*|X_i))}{\hat{G}(\tau^*)} \right]$$

(8)

where $V$ represents all censored and uncensored patients, $\delta_i = 1$ represents uncensored patients, $\hat{S}(\tau^*|X_i)$ is the predicted survival probability for patient $X_i$ at horizon $\tau^*$, $\hat{G}(\tau_i)$ is the probability of censoring (i.e. swapping censored and uncensored patients). Brier score for each imputed dataset was calculated separately, and then averaged to get pooled brier score.

### C-tau ($C\tau$)

$C\tau$ was calculated by considering patients survived longer than time $\tau$ as censored to examine the discriminative power of trained models for patients with different risk levels, as proved by [29].

### Area under receiver operator characteristics (AUROC)

AUROC was obtained by comparing predicted survival probabilities with binary outcomes of whether a patient survived longer than the prediction horizon, after excluding patients censored before the prediction horizon.

### Parkes's definition of serious error

Parkes' definition of serious error [30] is defined as predicted survival time is longer than 2 times of, or shorter than 0.5 times of actual survival time.

### Data analysis

The implementation of the GRU-D-Weibull model was performed on a unix system with NVIDIA Tesla V100 GPU. We reimplemented the original GRU-D code (in Keras format) associated with the original publication [22] to pytorch format (version 1.9.0) with python (version 3.8.10). The example code is available at github (https://github.com/xy-ruan/GRU-D-Weibull). Statistical analyses of AFT, XGB(AFT), and MTLR models were performed with R package survival (version 3.2.11), xgboost (version 1.6.0.1), MTLR (version 0.2.1) respectively.

### Data Availability

The datasets generated during and/or analyzed during the current study are not publicly available due to patient information confidentiality but are available from the corresponding author on reasonable request.

# Results

## Contribution of features to survival time by AFT model

Since the AFT model provides an explicit and easy-to-interpret estimate of the contribution and significance of each feature to patient survival time, we fitted the AFT model on the 5-fold dataset for each of the 3 imputations. A total of 15 models were fitted and the pooled results are shown in Table 3. Features significantly associated with longer survival time (i.e. a positive coefficient) are blood albumin and calcium. Those associated with shorter survival time (i.e. negative coefficient) are age, CHF, phosphorus, cirrhosis, CAD, smoking. Specifically, BMI, dyslipidemia, eGFR has only marginal to no significance after Bonferroni correction for multi-testing ($p * n (n = 19)$).

Table 3 Pooled correlation coefficients from AFT models

| Variable | Coefficient | Standard Error | P value |
|---|---|---|---|
| **Age** | -0.05 | 0.004 | **<0.001** |
| **CHF** | -0.921 | 0.079 | **<0.001** |
| **Albumin (Blood)** | 0.423 | 0.064 | **<0.001** |
| **Phosphorus** | -0.22 | 0.039 | **<0.001** |
| **Cirrhosis** | -0.638 | 0.118 | **<0.001** |
| **CAD** | -0.457 | 0.111 | **<0.001** |
| **Calcium** | 0.203 | 0.05 | **<0.001** |
| **Smoking** | -0.419 | 0.13 | **<0.001** |
| **BMI** | 0.017 | 0.005 | **0.002** |
| **Dyslipidemia** | 0.182 | 0.079 | **0.022** |
| **eGFR** | 0.007 | 0.004 | **0.044** |
| **uACR** | <0.001 | <0.001 | **0.048** |
| DM | -0.119 | 0.077 | 0.122 |
| Gender(Male) | -0.106 | 0.072 | 0.14 |
| SBP | 0.002 | 0.002 | 0.164 |
| Alcohol | -0.242 | 0.219 | 0.268 |
| Race(non-Caucasian) | 0.05 | 0.134 | 0.706 |
| DBP | -0.001 | 0.003 | 0.786 |
| Bicarbonate | -0.001 | 0.009 | 0.868 |

CHF: congestive heart failure CAD: chronic artery disease
BMI: body mass index eGFR: estimated glomerular filtration rate
DM: diabetes mellitus SBP: systolic blood pressure DBP: diastolic blood pressure

## C-index, $C\tau$, and AUROC

C-index measures a model's ability to correctly rank patients' order of events. At CKD4 index date, the four models have verY similar C-index (0.68, 95%CI (0.67,0.69) for AFT, XGB(AFT), and MTLR. 0.67~0.68 for GRU-D-Weibull depending on prediction horizon Fig 4 a). It is noteworthy that while both MTLR and GRU-D-Weibull do not assume proportional hazard, there are no remarkable differences across prediction horizons.

GRU-D-Weibull based C-indices improve throughout the follow-up time (Fig 4 a,b-1). The upward trend starts from 0.55 at ~3 years before CKD4 index date and ends at 0.67 on CKD4 index date for the 5-fold dataset, and reaches 0.77 at 4.3 years after CKD4 index date for the held-out dataset. Since C-index is implicitly dependent on time [29], we also analyzed $C\tau$ on 1,3,5 years prediction horizon, which indicates GRU-D-Weibull has better discrimination on 5-year prediction horizon before index date, and slightly better 1-year discrimination (than 3,5 years) after CKD4 index date. To explore the performance on predicting binary outcome (i.e. whether a patient develops an endpoint within a fixed horizon), we calculated AUROC for each prediction horizon (1,3,5 year(s)) separately by assuming absence of censoring (i.e. disregarding those censored before prediction horizon, and treating patients that survived beyond prediction horizon as survivor) (Fig 4 c). There is similar performance between GRU-D-Weibull and the AFT model on index date (AUROC=0.7~0.75 from 1 to 5 years). GRU-D-Weibull based AUROC has a remarkable upward trend throughout the follow-up time.

a)

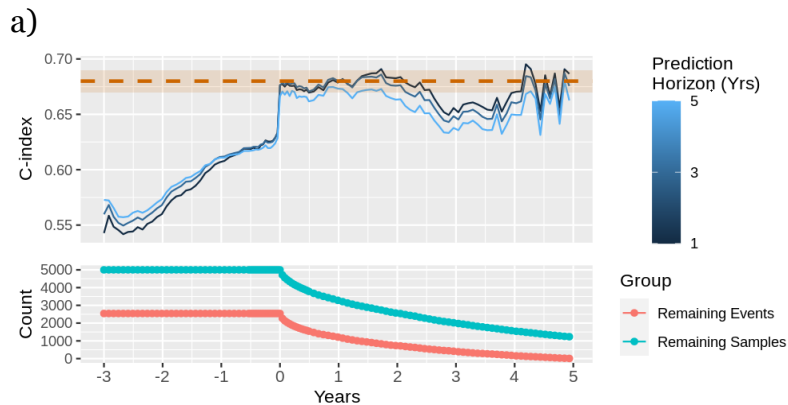

b)

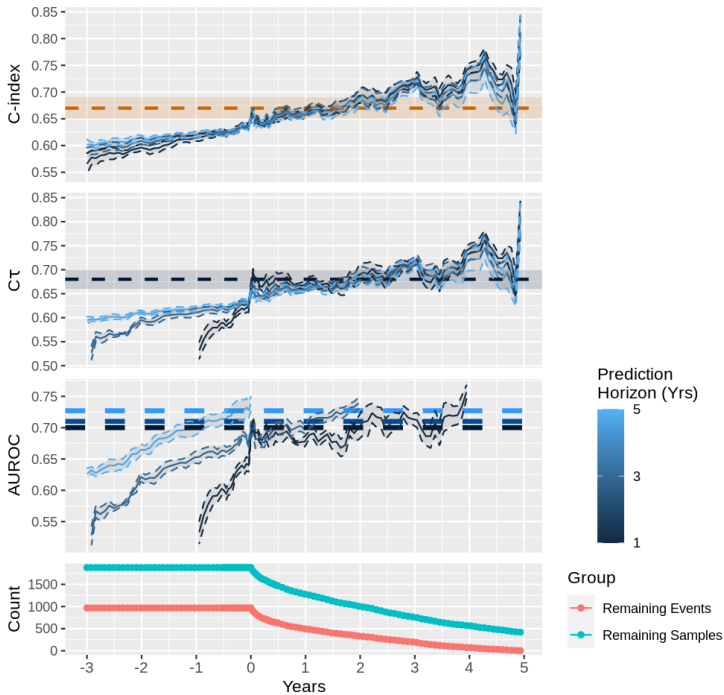

**Fig 4** GRU-D-Weibull based C-index across time points on 1,3,5 year(s) prediction horizon for **a)** 5-fold cross validation dataset, **b)** held-out dataset. Ribbons are 95%CI obtained by applying each of the 5 trained models to the held-out. **b-1** (C-index): dashed straight red line and ribbon mark the C-index and 95%CI of the AFT model (at CKD4 index date). **b-2** ($C\tau$): dashed straight blue line and ribbon mark the $C\tau$ with a 1 year horizon of the AFT model (very similar $C\tau$ for 3,5 years and not shown for clarity). **b-3** AUROC. Dashed straight blue lines mark the AUROC on the 1,3,5 year(s) prediction horizon of the AFT model. Note for GRU-D-Weibull the C-indices differ slightly across prediction horizons due to different shape parameter $\kappa$ for each patient, thus the constant proportional hazard assumption does not hold.

## L1-loss

L1-loss directly measures the difference between predicted and observed survival time. In this study, we used PMST as a point estimate of predicted survival time. GRU-D-Weibull has notably lower L1-loss and standard deviation (SD) than competing models in uncensored patients (i.e. developed endpoint during the 5-year follow-up). At CKD4 index date, the average absolute L1-loss is ~66% of XGB(AFT), ~60% of MTLR, and ~30% of AFT model. The SD is ~70% of XGB(AFT), ~75% of MTLR, and ~15% of AFT model (Supp Table 1). Throughout the follow-up time, GRU-D-Weibull shows improved performance (i.e. decreasing L1-loss) up to 3 years after CKD4 index date (Fig 5). The average absolute L1-loss is about 1 year after CKD4 index date.

a)

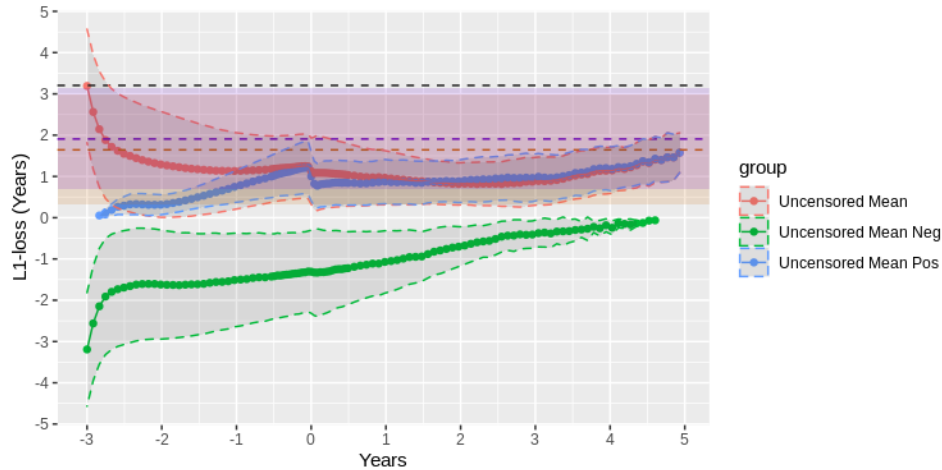

b)

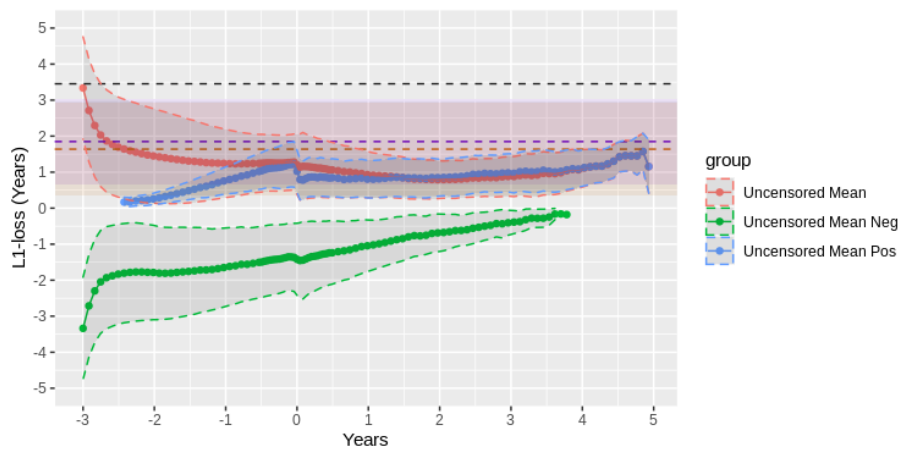

**Fig 5** L1-loss of uncensored patients based on differences between GRU-D-Weibull PMST and observed survival time (red curve) for a) 5-fold cross validation, b) held-out dataset. The dashed black, purple, and orange straight lines are mean prediction errors from AFT, MTLR, XGB(AFT) models, respectively, on CKD4 index date. Ribbons represent standard deviation (SD) (not shown for AFT for clarity). The blue and green curves and ribbons are L1-loss of GRU-D-Weibull based on only positive values (overestimate) and only negative values (underestimate), respectively. For the held-out dataset, data shown are average L1 margin loss and average SD by applying each of the 5 trained models to the held-out dataset.

At CKD4 index date, GRU-D-Weibull, XGB(AFT), and MTLR model have a similar proportion of ~60% predictions lie above Parkes' definition of serious error (i.e Target > 2*Prediction or Target < 0.5*Prediction) [30]. As a comparison, AFT has ~70% serious error. For GRU-D-Weibull, the error proportion decreases to ~40% about 2 years after CKD4 index date in both the 5-fold and held-out dataset (Fig 6). However, the error proportion increased remarkably after about 4 years of follow-up, mainly due to overestimation of survival time at the end of follow-up.

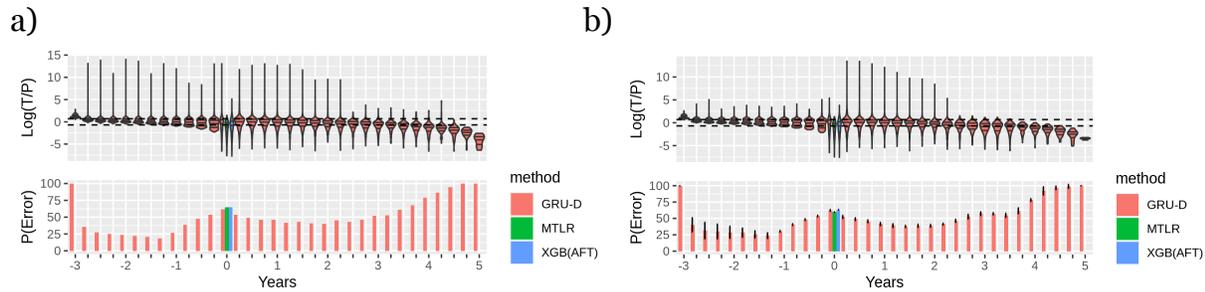

**Fig 6** Distribution of Log(Target/Prediction(*PMST*)) and proportion of serious error by Parkes' definition (i.e. Target > 2*Prediction or Target < 0.5*Prediction) for a) 5-fold cross validation, b) held-out dataset. Dashed horizontal black lines mark the location of $\pm log(2)$. For the held-out dataset, average predictions from 5 trained GRU-D-Weibull models are used for violin plot. The error bar represents the confidence interval of 5 trained GRU-D-Weibull models.

## 1-Calibration

1-Calibration measures the goodness-of-fit of predicted versus observed probability of event at specific prediction horizon, where a lower score represents better consistency. We splitted the patients into 10 chunks (a typical choice in survival study) ordered by predicted probability of survival at each of 1,3,5 year(s) prediction horizon, and calculated HL score by comparing with observed probability of event. At CKD4 index date, AFT model has the best calibration on 1,3,5 years horizons. Oppositely, XGB(AFT) has the worst calibration (Supp Table 1). GRU-D-Weibull has better (lower) HL score than MTLR model on 1 year horizon, and poorer calibration on 2-5 years horizons (Supp Table 1). GRU-D-Weibull has an overall decreasing HL score when approaching CKD4 index date, with a sudden increase around CKD4 index date (Fig 7) due to overestimation of the probability of event (Sup Fig 8). When the prediction horizon is fixed, there are similarly sloped linear relationships between predicted and observed probability of event at multiple follow-up time points (Sup Fig 8).

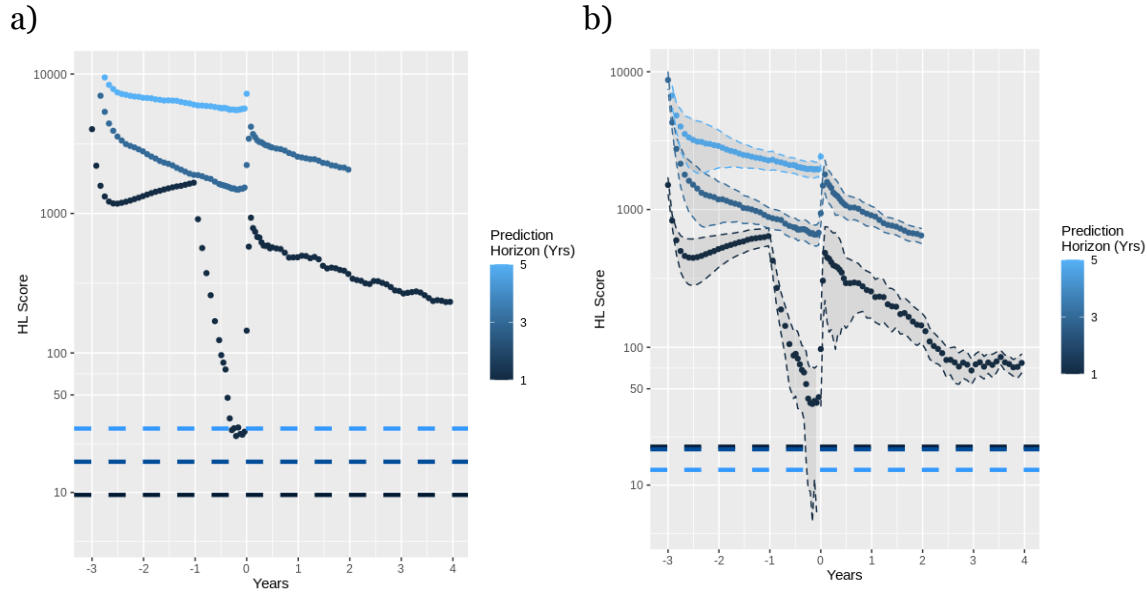

**Fig 7** GRU-D-Weibull based 1-Calibration on 1,3,5 years prediction horizon for a) 5-fold cross validation dataset, b) held-out dataset. Ribbons are 95%CI obtained by applying each of the 5 trained models to the held-out. Horizontal dashed lines mark the results from the AFT model on CKD4 index date. Notably, for the n year(s) prediction horizon, there is no patient reaching the endpoint until -n year(s) before index date.

**Brier Score**

Brier score measures both discrimination (i.e. correctly distinguish patients with/without event) and calibration of a model at specific prediction horizon, where a lower score represents better performance. At CKD4 index date, AFT and MTLR perform similarly on 1-5 years horizon, while XGB(AFT) is relatively poor on 1-4 years horizon. (Supp Table 1). GRU-D-Weibull has similar scores as AFT and MTLR at 1 year horizon, and better(lower) score at 5 years horizon in both 5-fold and held-out dataset (Supp Table 1). Throughout the follow-up time, GRU-D-Weibull has comparable or better Brier scores than all competing models (at CKD4 index date) on 1 year prediction horizon (Fig 8).

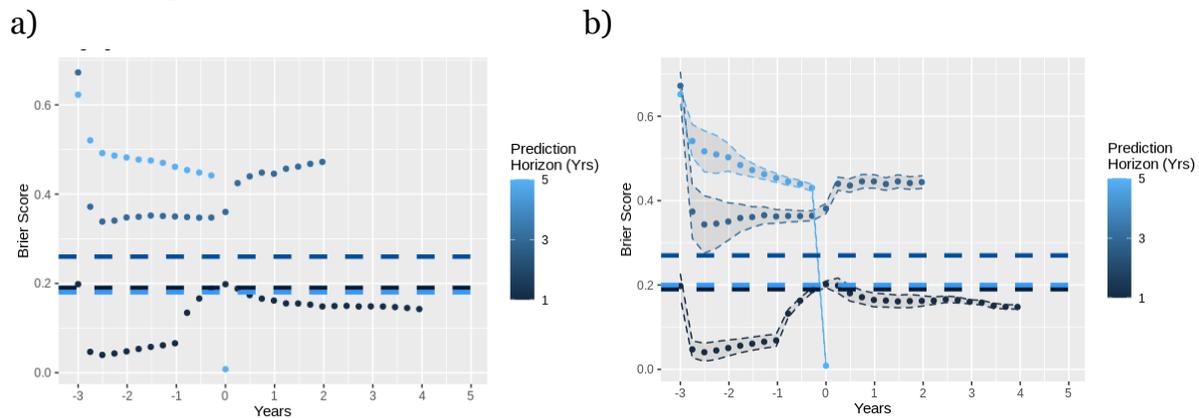

**Fig 8** GRU-D-Weibull based Brier Score on 1,3,5 years prediction horizon for a) 5-fold cross validation dataset, b) held-out dataset. Ribbons are confidence intervals obtained by applying each of the 5 trained models to the held-out. Horizon dashed lines mark results from the AFT model on CKD4 index date.

**Permutation feature importance across time**

Permutation tests across time indicate age, blood pressure (SBP, DBP), bicarbonate, CHF as the most important features that contribute to C-index (Fig 9). Among these features, age and bicarbonate have generally constant contributions across time. Similar to the result from the AFT model, SBP and DBP have little to zero impact on C-index on CKD4 index date. However, they become increasingly important throughout the follow-up time. CHF has a large contribution around index date but its impact decreases over time. As a comparison, the proportion of patients with CHF comorbidity (within 100-day window after CHF diagnosis) is around 30% both at CKD4 index date and at 3 years followup.

We see no contribution of eGFR throughout the entire follow-up period. Similarly, eGFR shows very little contribution in the AFT model. We noticed that at CKD stage 4 the EGFR is already at low level (<30) for the majority of patients (Sup Fig 3), thus its impact on survival time is hardly detectable without comparing to enough patients with normal to mild level (>60) eGFR.

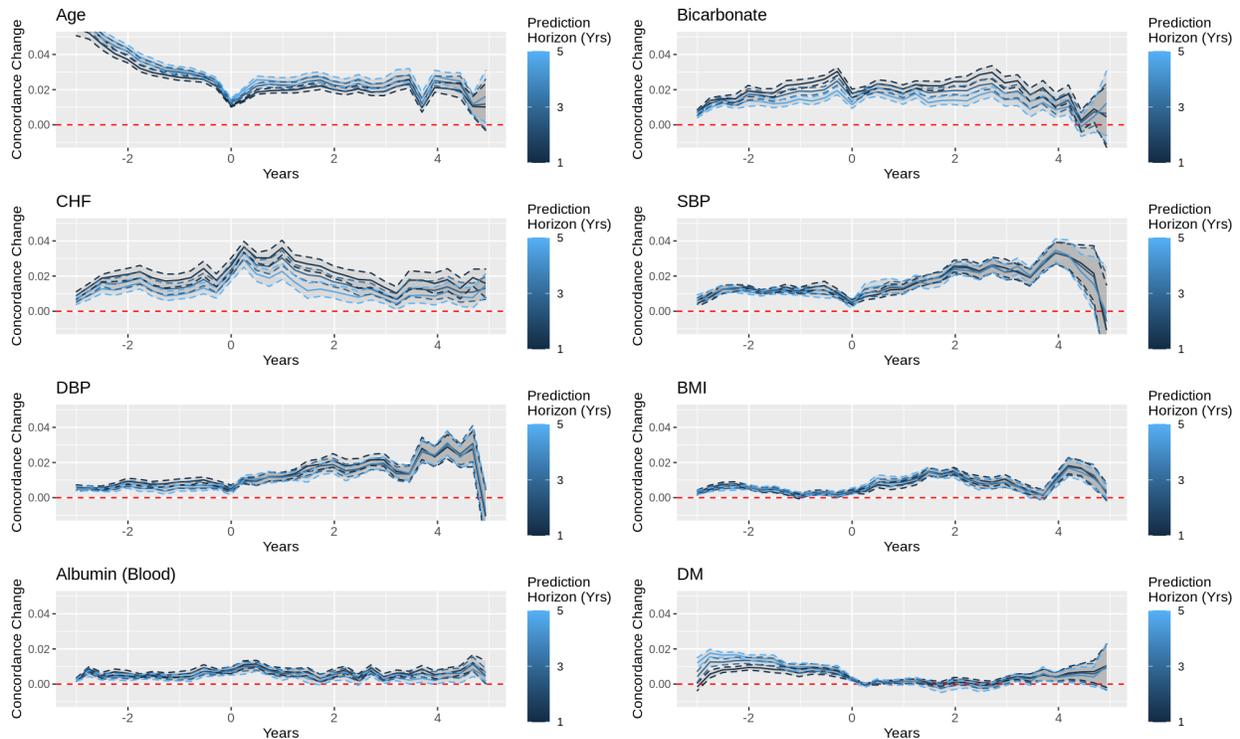

**Fig 9** Permutation feature importance across time for top 8 features. Y-axis represents change in C-index before and after permutation, which was performed for 1,3,5 year(s) prediction horizon.

**Partial dependence analysis**

Partial dependence analysis is considered as an important component of machine learning model explainability and is different from permutation feature importance in that it sheds light on the direction of effect (i.e. how changes in certain features increase/decrease predicted survival time). By shifting the feature of interest by a specific amount while keeping the rest of features untouched, we investigated how changes in specific single feature affect the prediction. Specifically, we still use PMST as target metric since it combines information from both the shape parameter $\kappa$ and scale parameter $\lambda$ and is a convenient point estimate of patient survival time. We examined eGFR and the 8 features with top importance in the permutation test. For each feature, we looked at PMST at different time points of follow-up (-2 to 4 years). We noticed that all continuous features except age have non-linear effects on PMST (Sup Fig 5). Specifically, increasing age linearly decreases PMST. Both SBP and DBP have ∩ (inverted U) shape effect, with best survival around normal level (~120mm Hg for SBP, ~60mm Hg for DBP) and deteriorate on either higher or lower pressure value. Bicarbonate has a similar ∩ shape effect as SBP and DBP with, however, a more remarkable decrease in survival time on high bicarbonate level. Higher eGFR increases survival time with, however, no remarkable change from 23 to 60 ml/min/1.73m2. Decreasing BMI (<19 kg/m2) remarkably decreases survival time whereas increasing

BMI to overweight (25-30 kg/m2) and non-severe obese (30-40 kg/m2) seem to increase survival time. Both increasing (from 3.6 to 4.8g/dL) and decreasing (from 3.6 to 2.7 g/dL) blood albumin increases predicted survival time. However, decreasing albumin below 2.7g/dL notably decreases survival time at CKD4 index date. Set CHF to yearly diagnosis notably decreases survival time. We see almost no effect of DM diagnosis on predicted survival time. Due to the slight overestimation of GRU-D-Weibull on survival time towards the end of follow-up time, the averaged PMST steps ¼ month higher yearly from CKD4 index date.

## Discussion
### Selection of comparison models
In this study, we compared GRU-D-Weibull architecture with AFT, XGB(AFT), and MTLR models on several major benchmark metrics related to individual survival distribution. Specifically, AFT was chosen because it is the most cited regression based survival model since first proposed by Cox DR[31] in 1972. XGB(AFT) represents the latest enhancement of the traditional AFT model through modern machine learning framework[24]. Finally, MTLR was chosen because it was shown to outperform Kaplan-Meier, AFT, COX-KP, COXEN-KP, and RSF-KM models on C-index, L1-loss, 1-Calibration, and Brier score under the majority of circumstances with varied sample and feature size, percentage of censoring, and follow-up time[27]. While we are aware of other survival modeling frameworks like DeepSurv[32], we believe current selections are sufficient for explorative and illustrative purposes.

### Advantages of GRU-D-Weibull architecture
The most notable properties of GRU-D-Weibull architecture are 1) improved concordance as more data become available throughout the follow-up time, and 2) remarkably more accurate point estimate of time-to-endpoint at individual level. Besides that, GRU-D-Weibull has several built-in advantages over competing models. First, missing values are parameterized asynchronously through built-in decay mechanisms so no imputation is required. Secondly, new measurements can be incorporated into the model at any time points to give an updated estimation of risk or time-to-endpoint. These properties are especially important for clinical application where measurements for various features may arrive interspersed throughout the follow-up. It is very hard, if not impossible, to train and maintain separate risk models to cater the needs of every patient at arbitrary time points.

MTLR performed similarly as AFT on C-index and Brier score, but poor on 1-Calibration. Thus for current analysis, the only advantage of MTLR is lower L1-loss on uncensored patients (50~60% of AFT), which is better (lower) than but directionally consistent with Haiders' results [27] on various datasets. We show that GRU-D-Weibull model based L1-loss on uncensored patients is ~60% of MTLR model at CKD4 index date, and continuously improves during follow-up. While the MTLR paper [25] mentioned it is possible to extend the MTLR model to update survival predictions with new measurement, the currently released package does not explicitly support real-time prediction, and if any, it is not a trivial task to properly incorporate the missing measurement.

Notably, the xgboost version of AFT model (XGB(AFT)) has even better (lower) L1-loss (achieved with *gbtree* booster but not *gblinear* booster, which makes unwarranted assumptions of linear association of features with the residuals after each epoch) than

MTLR model. It seems the improvement of XGB(AFT) in L1-loss is at a significant cost of performance on calibration. However, a carefully designed experiment is required to elucidate the trade-off between metrics.

According to [33], Parke's serious error "provides a realistic method of measuring prediction accuracy which can be acceptable in a wide variety of circumstances". While GRU-D-Weibull, XGB(AFT), and MTLR have a very similar serious error proportion (~60%) as doctor's prediction [34] on CKD4 index date, with currently limited input features GRU-D-Weibull is able to bring that down to ~40% during follow-up when new data become available. This constitutes a major step towards clinical utility of individualized endpoint prediction by turning the proportion of patients with serious prediction error from "majority" to "minority" [35]. Further improvement should be anticipatable when more input features are presented.

**Point estimate versus probability**
Most clinical prediction models describe a patient's likelihood of developing a certain disease as a risk score or probability at fixed predicted horizon [18,21]. While Haider [27,37] and Henderson [38] mentioned L1-loss as not a proper scoring rule, neither did they exemplify other more appropriate alternatives after brief mentioning of Log L1-loss[37], or exploring predictive intervals, categorical predictions, and relative risk methods [38]. A primary reason is the largely poor prediction accuracy, at least under traditional statistical frameworks. We emphasize L1-loss is arguably the most intuitive and challenging metric for measuring predictive accuracy. This is especially true for individual endpoint prediction, as both patients and healthcare providers would naturally expect the point estimate, if any, to be as accurate as possible. It is noteworthy that current GRU-D-Weibull architecture natively supports output of both point estimates and probability of survival up to arbitrary prediction horizon. The former is the most natural measure for patients. The latter can be handy for clinicians to formulate intervention plans to meet the rigid requirements of limited lifetimes of terminally ill patients.

**About Calibration**
AFT model has the best performance in 1-calibration, i.e. the predicted probability matches well with observed probability of survival. The main reason is that standard statistical tools, such as Cox's model, logistic regression, factorial designs largely belong to "population-based" models, where predictions are derived on distance from the population estimates. The current GRU-D-Weibull training configurations focus primarily on individual survival distribution. A direct evidence is the patient and timestep specific shape parameter $\kappa$. In another test (data not shown), we fixed $\kappa = 3.25 \pm 0.1$ (where modes equal to median survival time) and noticed remarkably improved 1-Calibration on 1 year prediction horizon, at the cost of other evaluation metrics. This emphasizes GRU-D-Weibull as a potential tool for facilitating individual level decision-making rather than for deriving population level statistics. Nevertheless, the consistently sloped linear relationship between predicted and observed probability of event across multiple follow-up points at fixed prediction horizon suggests the feasibility of post-training recalibration, which deserves further investigation.

Notably, the predicted probability of events is always higher than actual observation through the follow-up, which may be partially related to the medical interventions that changed the course of disease progression. This may also partly explain the sudden worsening of calibration around and after CKD4 index date, where intensive intervention should be anticipated.

**The choice of point estimate**
Besides of PMST, there are other point estimates like modes (i.e. the point maximizes Weibull PDF), predicted survival time $\lambda$, expected survival time $E(\tau)$ (i.e. $S(\tau)$ weighted sum of survival time), and minimum prediction error survival time (MPET) [33] that can be considered as a point estimate. We chose median survival time for several reasons. First, the modes could be undefined when $\kappa < 1$ according to its equation $\lambda(\frac{\kappa-1}{\kappa})^{1/\kappa}$. Secondly, using only $\lambda$ will discard the shape information. Thirdly, according to our test, $E(\tau)$ tends to more severely overestimate the survival time at the end of the follow-up period, and it is not part of our current loss function. Finally, MPET requires manually specifying an additional hyperparameter $k$ (the tolerance on fold difference between predicted and observed survival time), which adds another level of uncertainty to the results, and was shown to be less accurate than median survival time [36]. However, we do not rule out the utility of other measures as point estimates under other more appropriate circumstances.

**Feature importance and model explainability**
Except age, almost all continuous features have non-linear effects on predicted survival time. Specifically, certain features (e.g. DBP, SBP, and bicarbonate) have obvious parabola behavior, suggesting quadratic terms should be considered in traditional survival regression models. While age, DBP, SBP, CHF, and eGFR have direction of effect generally consistent with existing knowledge, the observations made on BMI, bicarbonate, blood albumin, and DM require special consideration. Our finding on BMI is echoed by [39], who analyzed ~0.45 Million US veteran and reported that low BMI (< 25 kg/m2) has worse outcome independent of CKD severity, while slightly overweight and mildly obese patients have the best outcome. For severe obese condition, we observed a flat region in averaged prediction when increase BMI to severe obese (40-45 kg/m2) range, while they documented slight increase in mortality risk with wide confidence interval in CKD4 patients. However, we should stay cautious when extrapolating beyond BMI 40 as both ours and [39]'s studies have limited numbers of patients in that range.

For bicarbonate, a common sense is that lower bicarbonate levels is a sign of metabolic acidosis [40], and maintaining bicarbonate level improves CKD outcomes [41]. However, [42] showed that high bicarbonate (> 24 mEq/L) in CKD are associated with poorer survival mainly due to increased heart failure, and similar results were observed in independent multi-ethnic study [43]. Our model suggests increasing bicarbonate level beyond 24 mEq/L has a more detrimental effect than decreasing bicarbonate level.

While decreasing blood albumin below 2.7 g/dL causes a remarkable decrease in predicted survival time is consistent with existing knowledge [44,45], we noticed an U-shape effect when blood albumin is between 2.7 and 4.8 g/dL, with lowest prediction at 3.6 g/dL, the lower boundary of normal range (3.4-5.4 g/dL). Since hypoalbuminemia itself is not a pathogenic factor [46], the hypoalbuminemia in current CKD4 patients should be considered as an indicator of the overall metabolic function and is confounded by other pathologic processes. As noted by [47] that the mortality rate associated with low blood albumin is dependent upon systematic inflammation in CKD5. Additionally, elevated serum albumin levels can be found in patients with dehydration and high protein diet consumption [48–50]. More study is required to elucidate how blood albumin is associated with the outcome of late stage CKD patients.

In both AFT model and GRU-D-Weibull feature importance and partial dependence test we see non-significant or little to no impact of DM diagnosis on either C-index or survival time, which is against common knowledge of diabetic kidney disease as a leading cause of ESRD [51]. Noteworthy is that in permutation test the we see the weak importance of DM diminished at CKD4 index date, leading us to hypothesize that rigorous blood sugar control of late stage CKD patients may offset the effect of DM, which however require medication data to be elucidated.

**Conclusion**
Individualized endpoint prediction is always an intriguing and challenging task but of great importance since it may offer great help in providing prompt and effective intervention. With built-in missing parameterization and capability of output both probability and point estimates for arbitrary prediction horizon throughout follow-up, GRU-D-Weibull is a potential candidate to meet the rigorous requirement of dealing with data sparsity, asynchronicity, individual uncertainty, and demand for real-time evaluation in real world clinical practice. Further research on its performance with more comprehensive input features, in-process or post-process calibration are warranted to benefit CKD4 or alike terminally-ill patients.

**Limitations**

Only a limited number of input features were considered in this explorative analysis. In reality, the survival of CKD patients could be affected by many other factors like medications (antihypertensive, antidiabetic drugs), medical interventions, change in lifestyle, hospitalization status, and other ICD comorbidities. A model with higher per-patient level prediction accuracy should be anticipated if these additional features were considered.

We used "Best Guess" survival time estimated from the MTLR model as a surrogate of the target time for censored patients, which makes the current training scheme less independent. There are several solutions with different complexity. The simplest one is using observed median survival time as a surrogate for censored patients, as mentioned by [52]. Another possibility is maximizing cumulative probability from time of censoring to upper bound, which is usually positive infinity for right censored data (i.e $CDF(+\infty) - CDF(\tau)$), as mentioned in [24]. Finally, it could be possible to start from a reasonable guess (e.g. median survival time of the whole dataset) for censoring patients, and iteratively updating the target time after each training epoch. Systematic investigation is required to benchmark each strategy and will be covered in future research.

There is an upward trend in GRU-D-Weibull PMST throughout follow-up time, from underestimation before to overestimation after CKD4 index date. Training separate models in different age groups only mildly alleviates the problem (data not shown). A possible reason is the model tries to minimize global error across the whole follow-up period, which makes the prediction worse on both ends. A plethora of techniques are available to tackle this issue, including pre-process input data (e.g. manipulate training target), in-process (e.g. additional error term to penalize under/over estimation), or post-process (e.g. making time point specific adjustment to $\kappa$ or $\lambda$ output to offset the error). We should also note that no assessment of fairness (i.e. comparable performance in different gender, race, age groups, etc.) is performed. A detailed solution is out of the scope of current paper as an explorative analysis of potential utility of GRU-D-Weibull architecture, and will be addressed in future research.

## Supplementary Figures

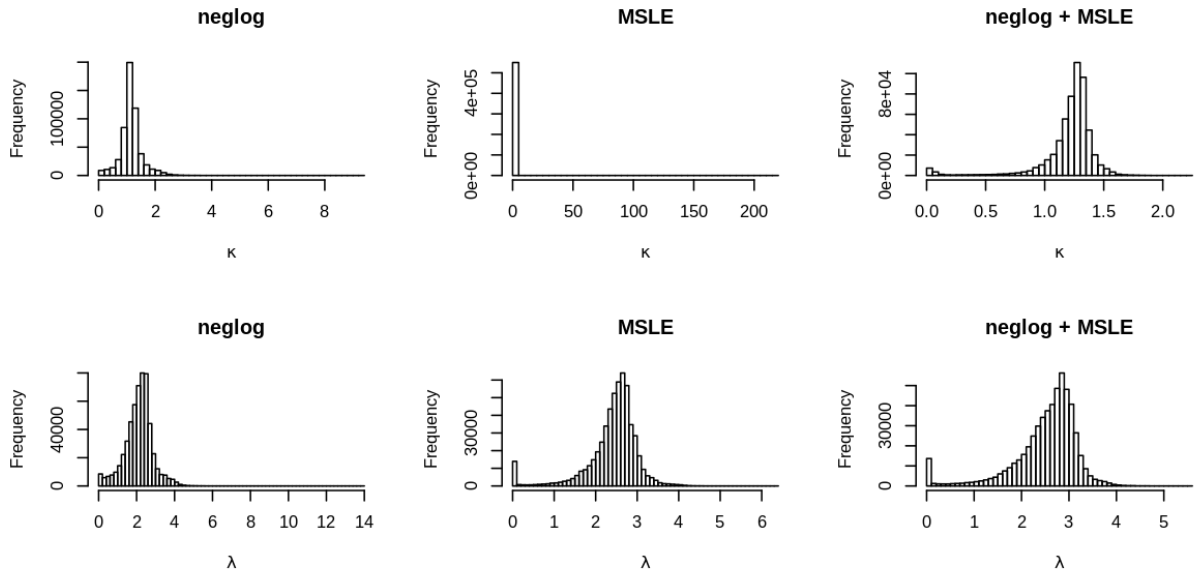

**Sup Fig 1** Distribution of shape parameter $\kappa$ and scale parameter $\lambda$ with *neglog* only, MSLE only, and composite loss function (*neglog* + MSLE).

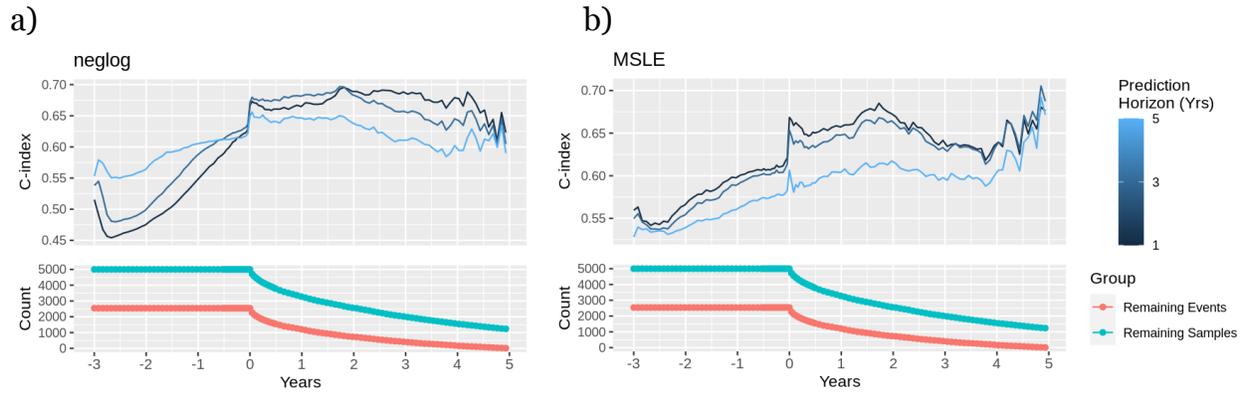

**Sup Fig 2** C-index throughout follow-up time with GRU-D-Weibull loss function composed of a) only *neglog*, b) only *MSLE*.

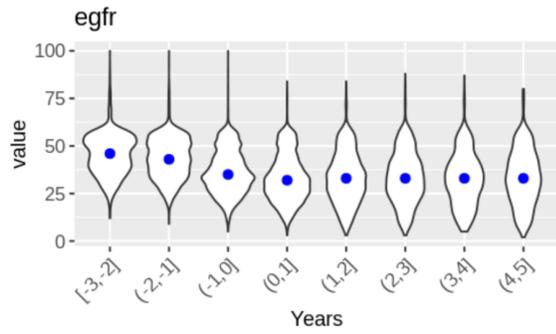

**Sup Fig 3**   Distribution of eGFR throughout follow-up time

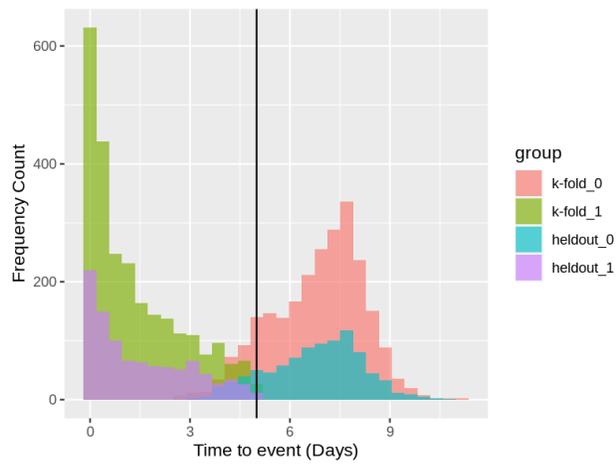

**Sup Fig 4**   Distribution of time to event at CKD4 index date. For uncensored patients ("k-fold_1", "heldout_1"), the observed time to event. For censored patients, the "Best Guess" time to event.

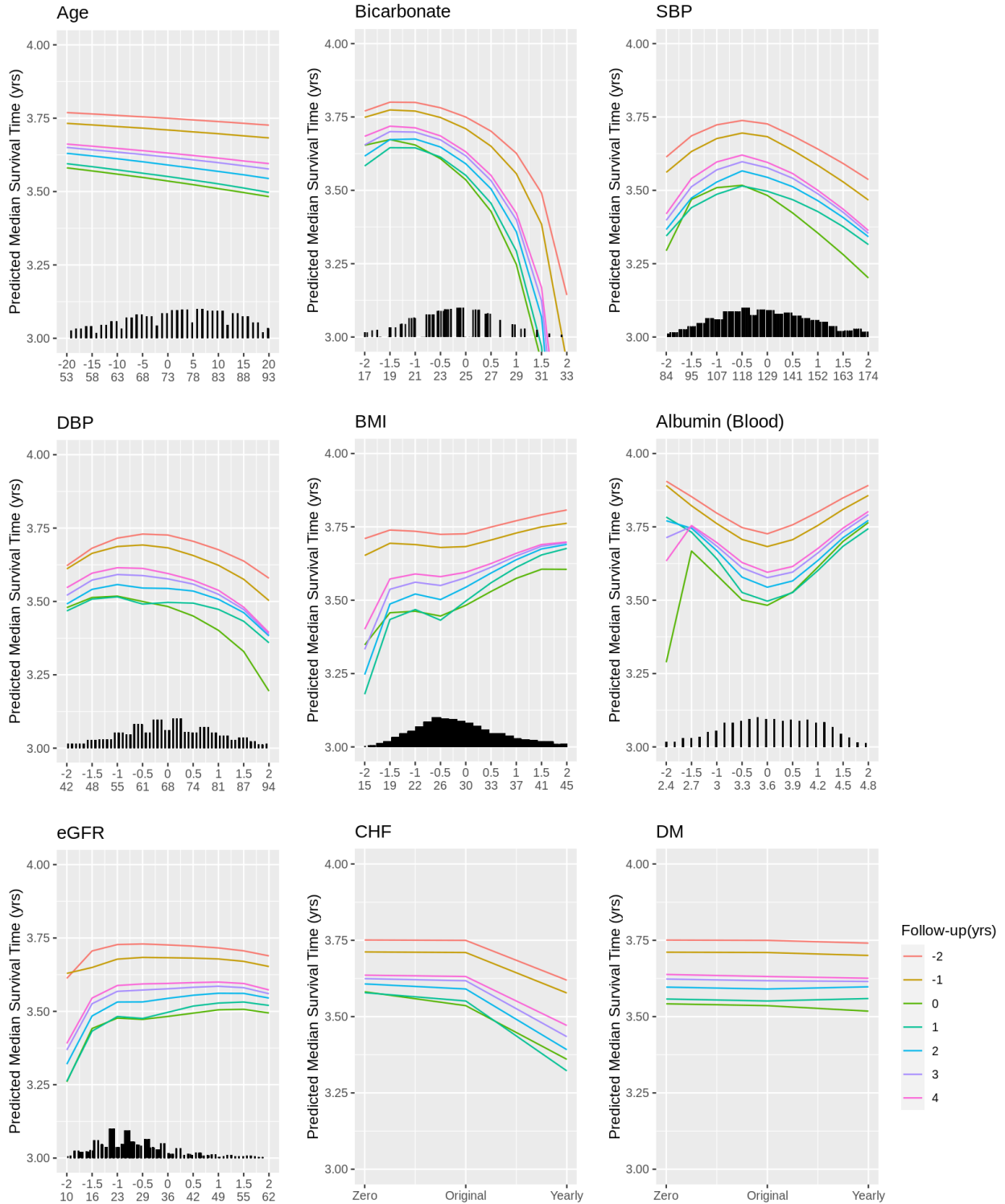

**Sup Fig 5** Partial dependence plot by shifting input value of each feature independently. Each line represents the median of PMST of the 1879 held-out patients averaged over 5-fold models. Continuous features except age are shifted by z-score. The second axis shows the mean value of the feature after shifting. The vertical bars illustrate the distribution of the corresponding feature in the training data around CKD4 index date. Categorical features are set to zero and yearly diagnosis, respectively. All plots have the same scale on Y-axis for easy comparison.

a)

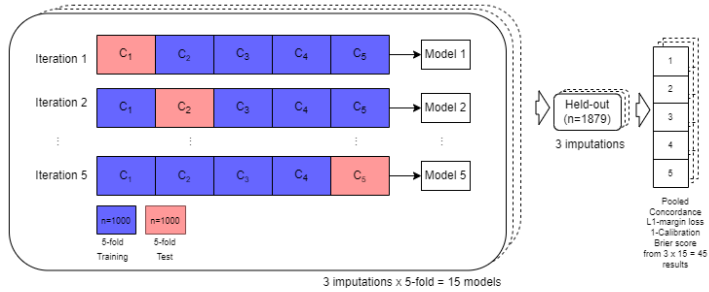

b)

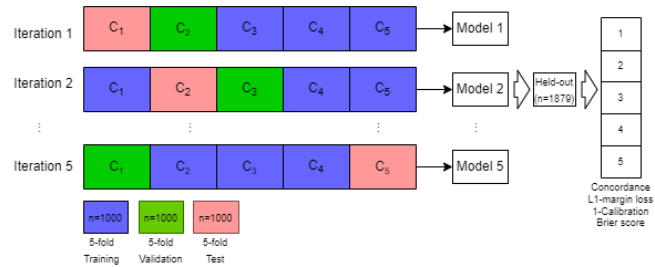

**Sup Fig 6** 5-fold cross validation and held-out strategy for a) AFT, MTLR model, and b) GRU-D-Weibull model. Specifically, XGB(AFT) uses an imputation scheme of a) and training scheme of b) since validation dataset is required for early stopping.

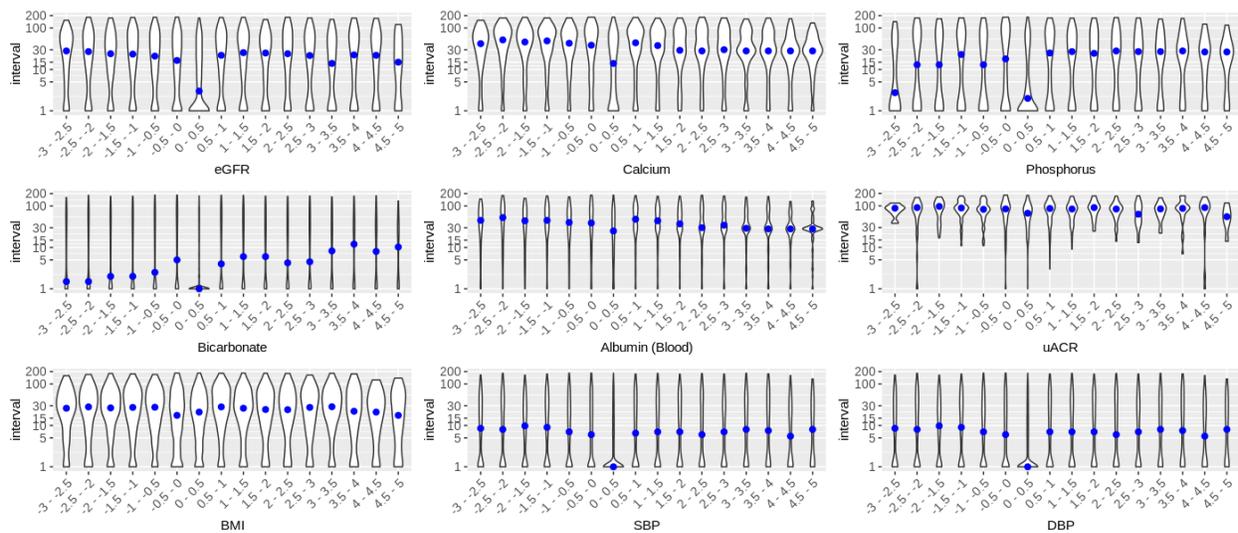

**Sup Fig 7**     Distribution of measurement time interval from 3 years before to 5 years after CKD4 diagnosis. Blue dots represent median value. **Same day measurements removed.**

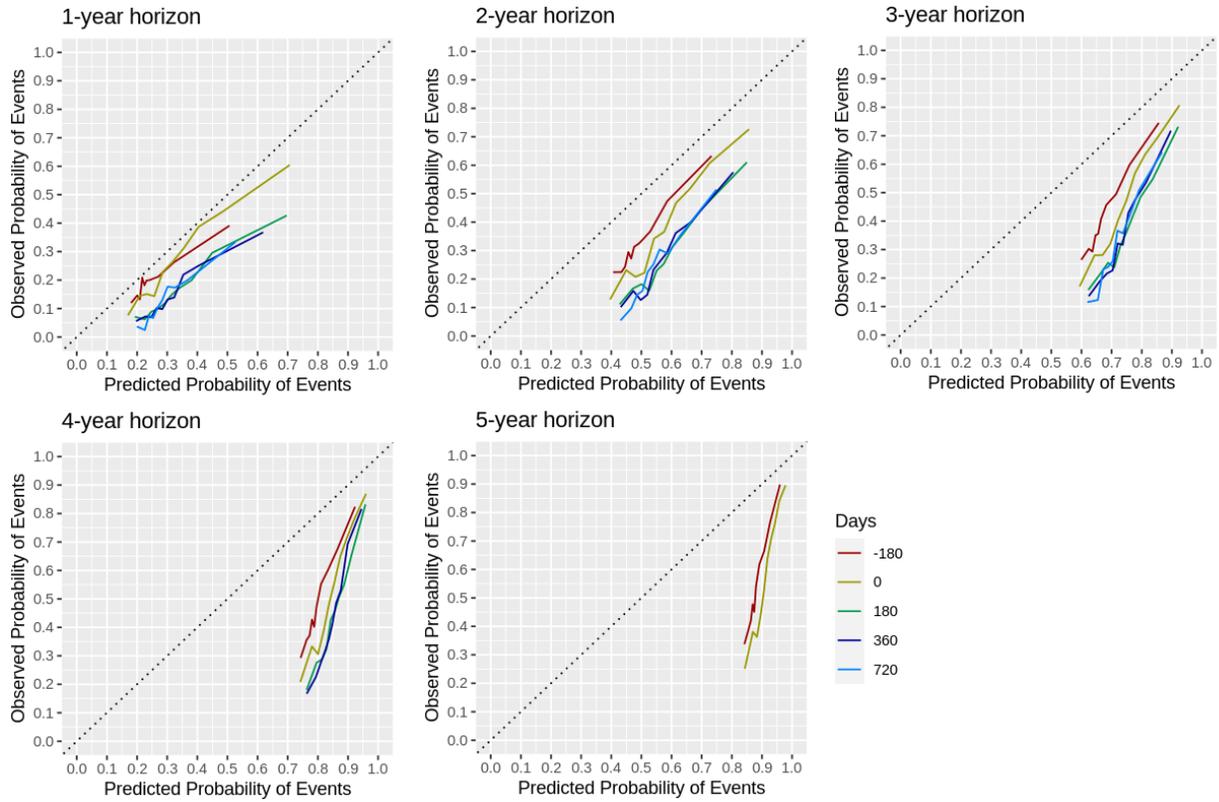

**Sup Fig 8**     Predicted versus observed probability of events on 1 to 5-year prediction horizon in the 5-fold cross validation dataset at different time points of follow-up.

| | 5-fold dataset | | | | Held-out | | | |
|---|---|---|---|---|---|---|---|---|
| Metrics | AFT | XGB(AFT) | MTLRa | GRU-D-Weibull b | AFT | XGB(AFT) | MTLRa | GRU-D-Weibull b |
| **L1 loss (mean, median, SD)** | | | | | | | | |
| uncensored(abs) | 1170, 629, 1970 | 600, 459, 487 | 696, 631, 448 | **394, 322, 319** | 1261, 660, 2831 | 598, 467, 471 | 675, 601, 435 | **419, 341, 357** |
| uncensored(pos) | 1317, 692, 2139 | 611, 459, 522 | 744, 680, 455 | **300, 278, 199** | 1394, 717, 3083 | 553, 395, 501 | 715, 650, 450 | **295, 268, 261** |
| uncensored(neg) | -411, -300, 346 | -401, -285, 355 | **-361, -279, 285** | -485, -396, 380 | -387, -277, 354 | **-337, -220, 329** | -359, -286, 283 | -531, -466, 383 |
| **1-calibration (HL, p-value)** | | | | | | | | |
| 1 year | **9.6, 0.292** | 414, <0.001 | 149, <0.001 | 26, <0.001 | **19.0, 0.015** | 174, <0.001 | 54.2, <0.001 | 40, <0.001 |
| 2 years | **13.2, 0.106** | 1332, <0.001 | 159, <0.001 | 480, <0.001 | **28.2, <0.001** | 641, <0.001 | 68.4, <0.001 | 267, <0.001 |
| 3 years | **16.6, 0.035** | 3325, <0.001 | 172, <0.001 | 1504, <0.001 | **18.2, 0.020** | 1540, <0.001 | 61.5, <0.001 | 658, <0.001 |
| 4 years | **15.2, 0.055** | 7035, <0.001 | 206, <0.001 | 3134, <0.001 | **10.0, 0.267** | 2534, <0.001 | 58.3, <0.001 | 1125, <0.001 |
| 5 years | **28.7, <0.001** | 16049, <0.001 | 285, <0.001 | 5621, <0.001 | **12.9, 0.117** | 4993, <0.001 | 81.3, <0.001 | 1964, <0.001 |
| **Brier score** | | | | | | | | |
| 1 year | 0.19 | 0.21 | 0.2 | 0.2 | 0.19 | 0.2 | 0.19 | 0.2 |
| 2 years | 0.24 | 0.25 | 0.24 | 0.28 | 0.24 | 0.26 | 0.24 | 0.3 |
| 3 years | 0.26 | 0.28 | 0.27 | 0.36 | 0.27 | 0.3 | 0.27 | 0.38 |
| 4 years | 0.27 | 0.29 | 0.28 | 0.42 | 0.28 | 0.3 | 0.29 | 0.42 |
| 5 years | 0.18 | 0.09 | 0.16 | 0.008 | 0.2 | 0.09 | 0.17 | 0.009 |

a For MTLR and GRU-D-Weibull, C-indices are shown as range (minimum and maximum 95% CI) for 1-5 years prediction horizon.

b Based on best results within 30 days around index date